\documentclass[11pt,a4paper]{article}
\usepackage[hyperref]{eacl2021}
\usepackage{times}
\usepackage{latexsym}
\usepackage{cancel}
\usepackage{graphicx}
\usepackage{amsmath}
\usepackage{latexsym}
\usepackage{amssymb}
\usepackage{amsthm}
\usepackage{color}
\usepackage{xspace}
\usepackage{subfigure}
\usepackage{array, booktabs, makecell}

\usepackage{booktabs}
\usepackage{array,tabularx}
\usepackage{ulem}
\usepackage{amssymb}
\usepackage{pifont}
\usepackage{multirow}

\newenvironment{conditions*}
  {\par\vspace{\abovedisplayskip}\noindent
   \tabularx{\columnwidth}{>{$}l<{$} @{}>{${}}c<{{}$}@{} >{\raggedright\arraybackslash}X}}
  {\endtabularx\par\vspace{\belowdisplayskip}}

\usepackage{microtype}

\aclfinalcopy 

\title{EIGEN: \textbf{E}vent \textbf{I}nfluence \textbf{GEN}eration \\ using Pre-trained Language Models}

\author{Aman Madaan~\thanks{\hspace{0.5em} authors contributed equally to this work.}\hspace{0.5em}, Dheeraj Rajagopal~\footnotemark[1]\hspace{0.5em}, Yiming Yang,\\
\textbf{Abhilasha Ravichander, Eduard Hovy, Shrimai Prabhumoye}  \\
  Language Technologies Institute, Carnegie Mellon University \\
  Pittsburgh, PA, USA \\
  \texttt{\{amadaan,dheeraj,yiming,aravicha,ehovy,sprabhum\}@cs.cmu.edu} \\}

\date{}

\begin{document}
\definecolor{Red}{rgb}{1,0,0}
\definecolor{Green}{rgb}{0,0.7,0}
\definecolor{Blue}{rgb}{0,0,1}
\definecolor{Red}{rgb}{0.6,0,0}
\definecolor{Orange}{rgb}{1,0.5,0}
\definecolor{yellow}{rgb}{0.65,0.6,0}
\definecolor{cadmiumgreen}{rgb}{0.0, 0.42, 0.24}

\newcommand{\am}[1]{\textcolor{Green}{[#1 \textsc{--Aman}]}}
\newcommand{\dheeraj}[1]{\textcolor{Blue}{[#1 \textsc{--Dheeraj}]}}
\newcommand{\sm}[1]{\textcolor{Orange}{[#1 \textsc{--SM}]}}
\newcommand{\notes}[1]{\textcolor{Red}{[#1 \textsc{--notes}]}}
\newcommand{\ed}[1]{\textcolor{cyan}{[#1 \textsc{--Ed}]}}
\newcommand{\lasha}[1]{\textcolor{yellow}{[#1 \textsc{--Abhilasha}]}}
\newcommand{\ar}[1]{\textcolor{red}{\bf\small [#1 --Abhilasha]}}
\newcommand{\bhavana}[1]{\textcolor{red}{\bf\small [#1 --bhavana]}}
\newcommand{\secref}[1]{\S\ref{#1}}
\newcommand\given[1][]{\:#1\vert\:}

\newcommand{\numgraphs}{2107\xspace}
\newcommand{\numpassages}{379\xspace}
\newcommand{\numquestions}{40.7K\xspace}

\newcommand{\lrate}{\textcolor{Red}{LR-HERE} }
\newcommand{\dropout}{\textcolor{Red}{DROPOUT-HERE} }
\newcommand{\red}[1]{\textcolor{red}{#1}}
\newcommand{\green}[1]{\textcolor{green}{#1}}
\newcommand{\cadmiumgreen}[1]{\textcolor{cadmiumgreen}{#1}}

\newcommand{\helps}{\overset{+}{\longrightarrow}}
\newcommand{\hurts}{\overset{-}{\longrightarrow}}
\newcommand{\helpedby}{\overset{+}{\longleftarrow}}
\newcommand{\hurtby}{\overset{-}{\longleftarrow}}
\newcommand{\relatedby}{\overset{r}{\longrightarrow}}
\newcommand{\ir}{\textsc{ir}\xspace}
\newcommand{\wiqa}{\textsc{wiqa}\xspace}
\newcommand{\bert}{\textsc{bert}\xspace}
\newcommand{\qa}{\textsc{qa}\xspace}
\newcommand{\ours}{\textsc{eigen}\xspace}
\newcommand{\comet}{\textsc{comet}\xspace}
\newcommand{\plm}{\textsc{plm}\xspace}
\newcommand{\cmark}{\ding{51}}%
\newcommand{\xmark}{\ding{55}}%
\newcommand{\lms}{\textsc{lm}\xspace}

\maketitle
\begin{abstract}

Reasoning about events and tracking their influences is fundamental to understanding processes.
In this paper, we present \ours~- a method to leverage pre-trained language models to generate event influences conditioned on a context, nature of their influence, and the distance in a reasoning chain.
We also derive a new dataset for research and evaluation of methods for event influence generation.
\ours outperforms strong baselines both in terms of automated evaluation metrics (by 10 \textsc{rouge} points) and human judgments on closeness to reference and relevance of generations.
Furthermore, we show that the event influences generated by \ours improve the performance on a ``what-if'' Question Answering (\wiqa) benchmark (over 3\% F1), especially for questions that require background knowledge and multi-hop reasoning.
\end{abstract}

\newcommand{\yy}[1]{\textcolor{blue}{\bf\small [#1 --YY]}}

\normalem
\section{Introduction}
\begin{figure*}[!t]
    {\includegraphics[width=\textwidth]{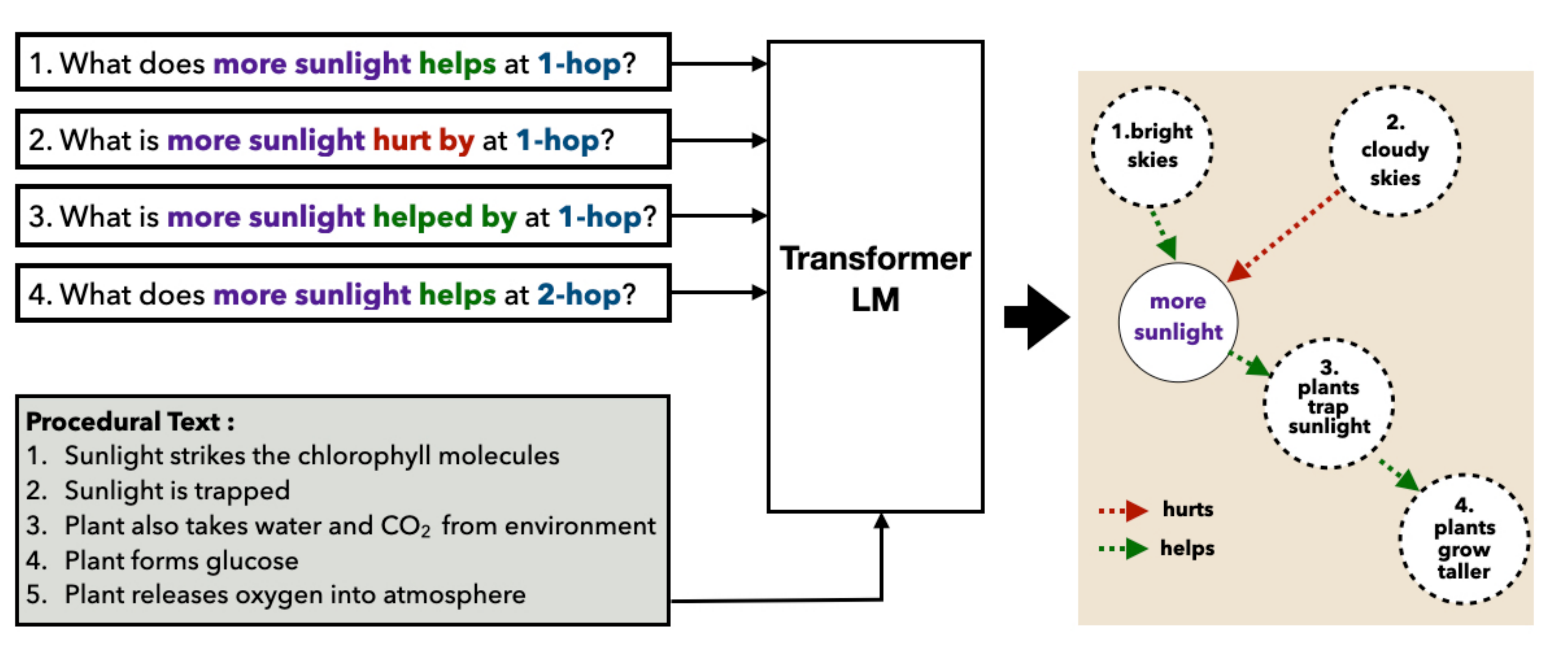}}
    \caption{An overview of our methodology. The procedural text describes the process of \emph{photosynthesis}. For this example, we generate the influence graph for the event \emph{more sunlight}. The influence graph is generated for the relation types - \emph{helps, hurt by, helped by} and hops = $\{1,2\}$. A sample output influence graph shows the generated events - \emph{bright skies, cloudy skies, plants trap sunlight, and plants grow taller}} 
    \label{fig:intro-example}
\end{figure*}
Humans are adept at anticipating and reasoning about events and their causal effects~(influences) on other events.
Consider these questions - \emph{Would it rain more if we plant more trees?},  \emph{What would help the water in boiling faster?} - answering these questions requires the ability to comprehend the complex processes of \emph{plant growth} and \emph{water boiling} and the capacity to reason about how various events influence each other in these processes that are typically implicit in text. Hence, reasoning about events and influences remains a significant challenge for machines.
Understanding such events and tracing their influence chains is essential for end tasks like question answering (\qa)~\cite{tandon2019wiqa}, process tracking~\cite{dalvi-etal-2018-tracking}, reasoning about qualitative relationships~\cite{tafjord2019quarel}, and physical commonsense reasoning~\cite{sap2019atomic,bisk2020piqa}.


Previous approaches have studied event understanding in the context of event extraction \cite{chambers-jurafsky-2008-unsupervised,Yang2019ExploringPL,wang-etal-2019-open}, temporal event reasoning~\cite{ning2018joint,vashishtha2019fine}, and Question-Answering~\cite{tandon2019wiqa,dalvi-etal-2018-tracking}.
However, these systems are primarily \emph{extractive} --- they reason about events already mentioned in the text, limiting their ability to be integrated to downstream tasks that require implicit reasoning about events. 
The task of generating novel event influence in unseen contexts is still an open challenge. 

Meanwhile, promising evidence from recent work attests to the ability of pretrained language models (\plm) to encode a wide-range of knowledge from their pretraining corpus \cite{Bosselut2019COMETCT,petroni2019language,davison-etal-2019-commonsense}, enabling their successful adaptation in downstream tasks \cite{yang2020g,kumar2020data,Guan2020AKP}.  
Motivated by these successes, we investigate whether we can adapt \plm for the novel task of \emph{event influence generation} and determine empirically whether the generated event influences lead to downstream performance gains.
Such an exploration entails two major challenges: i) lack of large-scale stand-alone datasets to study event influences, and ii) a framework to leverage \plm to adapt them for event influence generation. 

In this work, we address these challenges by first deriving a large corpus based on WIQA~\cite{tandon2019wiqa} dataset that can be used for the generation of event influences conditioned on context, relationship between the events, and the distance between them in a reasoning chain.
Next, we propose our framework, \ours, that takes a context and an event, and generates its influences both in forward and backward directions.
An example use of our framework is shown in Figure \ref{fig:intro-example}.
In the figure, nodes represent the event influences and the edges represent the nature of the influence (relation) between them. 
These relations can either be positive (when one event \textit{helps} the occurrence of another) or negative (when one event \textit{hurts} the occurrence of another). 
The distance between any given pair of nodes (in terms of number of edges traversed) is denoted by \emph{hop}.

\ours fine-tunes a \plm to generate novel event influences for unseen contexts using masked language modeling.
We show empirically that our framework generates high quality influences for an event, both in terms of automated metrics (by $\sim$ 10 ROUGE) and human metrics --- relevance and proximity to the reference text. 
Together, the overall framework can be seamlessly integrated into any downstream task.
In one such instance, we show how the event influences generated from \ours can be easily augmented to a downstream \qa task and improve its performance without any need for modifying the underlying model architecture. 
In summary, our contributions are:
\begin{enumerate}
     \item We propose the task of \emph{event influence generation} and derive a large-scale dataset for the same. 
  
    \item We propose \ours, a framework to generate targeted influence nodes for an event. Our experiments show that \ours outperforms strong baselines in both automated and human evaluation. 

    \item We also validate our approach by augmenting generated influences to a downstream \qa dataset, improving over the state of the art by 3\% in overall accuracy, and by 8\% on the subset of questions that require implicit event-influence reasoning \footnote{Code and data available at \url{https://github.com/dheerajrajagopal/EIGEN}.}.
\end{enumerate}
    
\section{Related Work}
\label{sec:rel_wrk}

\paragraph{Event Influences:} There has been immense interest in understanding event chains in stories and news corpora in both unsupervised \cite{chambers-jurafsky-2008-unsupervised}  and supervised \cite{rudinger2015script,liu-etal-2018-narrative} settings. 
Such approaches aim to extract the event chains that are explicitly mentioned in the input text and are unyielding towards implicit event reasoning.
Events and their influences have also been studied in restricted domains such as cooking recipes~\citep{kiddon2016globally, bosselut2018simulating}, and in general procedural text~\citep{dalvi-etal-2018-tracking} as a classification task over a restricted set of events.
\citet{tandon2019wiqa} introduce the \wiqa dataset, which relaxes this restriction by collecting event perturbations over general procedures, where the goal is to predict the influence between two given events (positive, negative or no-effect), while also providing explicit annotations for capturing the influences over multiple reasoning hops. 
Albeit being resourceful, restricted task formulation limits use of these datasets to adapt for event influence generation task. To overcome this challenge, we derive a large-scale event-influence dataset from \wiqa (discussed in Section~\secref{sec:approach}).


\paragraph{Language Models for Knowledge Generation:}
The use of large scale neural networks to generate knowledge has been studied under various task settings.
\citet{sap2019atomic} use LSTM-based encoder-decoder architectures to generate general-purpose social commonsense knowledge.
These methods were then improved by replacing LSTMs with large-scale pre-trained transformer language models.
\citet{Bosselut2019COMETCT} proposed \comet, which fine-tunes \textsc{gpt}~\citep{radford2018improving} on \textsc{atomic}~\cite{sap2019atomic} and conceptnet~\citep{speer2017conceptnet} for knowledge-completion task.
An extension to this work by incorporating structural and semantic constraints was proposed by \citet{malaviya2019exploiting}.
Similar to \citet{Bosselut2019COMETCT}, we leverage pre-trained language models for the conditional generation of events.
However, unlike \comet, we i) condition our generations on a larger context and hop-information, and ii) provide a framework for recursively generating event influence graphs for a given process and an event.
Additionally, unlike \comet, a dataset that can be used for our task is not readily available, and hence we outline a method for adapting existing datasets for our task as an additional contribution. 
\section{Event Influence Generation}
\label{sec:approach}

\ours is a framework for generating fine-grained event influences for a given context, conditioned on the relation and the hop information.
\ours leverages a pretrained language model to learn to generate novel event influences over multiple hops. 
In this section, we present (i) our task formulation (section \secref{sec:task}), (ii) the dataset collection process (section \secref{sec:dataset}) and (iii) the learning procedure (\secref{sec:masked-lm}). 


\subsection{Task}
\label{sec:task}
We formalize the event influence generation task as follows:
Given an input tuple $(P, n_s, r, h)$, where \\
(i) $P$ is a procedural passage $P$ that describes the steps in a process, \\
(ii) $n_s$ is an event in $P$ for which the influences are to be generated, \\
(iii) $r$ is an influence relation that describes the nature of the influence and, \\
(iv) $h$ is the hop length (distance) between the event $n_s$ and its influence in a reasoning chain, our task is to generate a target event $n_t$ such that $n_s \relatedby n_t$ at hop $h$ in the context of $P$.
We focus on 4 broad classes of event influence relations $r$ between events $n_s$ and $n_t$:  \\
(1) \textit{helps:} $n_s$ positively influences $n_t$ ($n_s \helps n_t$) \\
(2) \textit{hurts:} $n_s$ negatively influences $n_t$ ($n_s \hurts n_t$) \\
(3) \textit{helped-by:} $n_s$ is positively influenced by $n_t$ ($n_s \helpedby n_t$), and \\
(4) \textit{hurt-by:} $n_s$ is negatively influenced by $n_t$ ($n_s \hurtby n_t$).

We show an example of our task in Figure~\ref{fig:intro-example}, where we generate the influences for the event \textit{more sunlight }in the context of \textit{photosynthesis}.
In this example, for the relation \emph{hurt-by}, and a hop-length $h=1$, we aim to generate the text $\textit{cloudy skies}$ ($n_t$).
Similarly, given $h=2$ and a relation \emph{helps}, the system generates the target event $\textit{plants grow taller}$.
Note that the generation of a node refers to the generation of text tokens describing the node.

\subsection{Dataset}
\label{sec:dataset}
Lack of datasets remains a challenge for studying the task of event influence generation.
To address this challenge, we adapt \wiqa~ \cite{tandon2019wiqa} to generate a large-scale event influence generation dataset.
\wiqa consists of a set of procedural passages, each accompanied by a human-curated \textit{influence graph}.
The influence graph captures the interactions between the events and their influences and external perturbations in the context of the process described by the passage. Although these graphs can be subjective, \wiqa has high inter-annotator agreement\footnote{0.6 Krippendorff's alpha}, motivating our choice to leverage these graphs. 

We decompose the influence graphs to create our generation dataset. 
An influence graph for a passage $P$ is denoted by by $G=(V,E)$, where $V$ denotes the set of vertices and $E$ the set of edges. 
The nodes $n \in V$ represent the events, and the edges represent the relationship (\textit{helps} or \textit{hurts}) between them.
Each edge $n_s \relatedby n_t \in G$ contributes a sample for our training data, composed of tuples of the form $\mathbf{x}_i = (P, n_s, r, h)$ and $\mathbf{y}_i = n_t$.

For creating multi-hop training samples for our task, we exploit the transitive compositionality of the influence relations.
For example, if $(n_a \helps n_b) \land  (n_b \helps n_c) \equiv (n_a \helps n_c)$.
Similarly, $(n_a \helps n_b) \land (n_b \hurts n_c) \equiv (n_a \hurts n_c)$.
In summary, $n_s \hurts n_t$ if the path from $n_s$ to $n_t$ has an odd number of $\textit{hurts}$ edges, and $n_s \helps n_t$ otherwise.
For example,  $\textit{cloudy skies} \hurts \textit{plants grow taller}$ in Figure~\ref{fig:intro-example}.

We also augment the dataset with inverse influences, where our goal is to capture event influence in the reverse direction. 
For example, if $n_a \helps n_b$, then $n_b \helpedby n_a$.
After augmentation, our dataset captures diverse influences with respect to the relations and hops as described in section \secref{sec:task}. A detailed dataset statistic is shown in Table \ref{table:data-split}. 

\begin{table}[!h]
 \resizebox{\columnwidth}{!}{
\begin{tabular}{llrrrr}
\toprule
Split & Relation Type & 1-Hop & 2-Hop & 3-Hop & Total\\ \midrule
train & helps         & 8723 & 13085  & 5815 & \multirow{4}{*}{119.2k} \\
train & hurts         & 13081 & 13088 & 5815  \\
train & is helped by  & 8723 & 13085  & 5815  \\
train & is hurt by    & 13081 & 13088 & 5815  \\ \midrule
test  & helps         & 1382  & 2075  & 922  & \multirow{4}{*}{18.8k}  \\
test  & hurts         & 2073  & 2075  & 922   \\
test  & is helped by  & 1382  & 2075  & 922 \\
test  & is hurt by    & 2073  & 2075  & 922   \\ \midrule
dev   & helps         & 2547  & 3824  & 1697  & \multirow{4}{*}{34.8k} \\
dev   & hurts         & 3824  & 3823  & 1697  \\
dev   & is helped by  & 2547  & 3824  & 1697  \\
dev   & is hurt by    & 3824  & 3823  & 1697  \\ 
\bottomrule
\end{tabular}
}
\caption{Breakdown of number of samples by relation type, distance, and split. We maintain the same train-dev-test split as the \wiqa~ dataset.}
\label{table:data-split}

\end{table}

Although our dataset uses relationship types from Table \ref{table:data-split}, our framework makes no relation-specific assumptions and is generally applicable to a broader range of relationships.

\begin{table*}[t]
\centering
\begin{tabular}{lrrrrrrr}
\toprule
Model   & BLEU-1 & BLEU-2 & BLEU-3 & BLEU-4 & METEOR & ROUGE & Polarity \\\midrule
\textsc{gpt-2} w/o Fine-tuning* &  7.66   &  3.05      &  1.56      & 0.91      &  4.79  & 7.85 & 8.25  \\   
\textsc{lstm} Seq-to-Seq &  17.65 & 	7.51 & 	5.16 & 	4.26 & 	7.69 & 	18.71 & 	70.22   \\
\comet &   20.63   & 10.01 & 7.12  &  5.93 & 8.82 & 20.93  & 71.97  \\
\textbf{\ours} &   \textbf{28.97}   &  \textbf{16.23} & \textbf{11.69}  &  \textbf{9.74} & \textbf{12.85} & \textbf{29.65}  & \textbf{77.24} \\
\bottomrule
\end{tabular}
\caption{Generation Quality for \ours and the baseline. BLEU-n refers to geometric average of BLEU scores calculated upto n-grams. *- indicates that this baseline model is not fine-tuned on our dataset.}
\label{tab:gen-quality}
\end{table*}

\subsection{Learning to Generate Influences}
\label{sec:masked-lm}

As discussed in section~\secref{sec:task}, the training data consist of samples $(\mathbf{x}_i, \mathbf{y}_i)$, where $\mathbf{x}_i = (P_i, n_s, r_i, h_i)$ and $\mathbf{y}_i$ is the corresponding target node $n_t$.
In our dataset, each procedural passage is used to create multiple training examples from variations in $n_s, r_i, h_i$.
For instance, Figure~\ref{fig:intro-example} shows four such training samples, where one example is as follows:
$\mathbf{x}_i = (P_i = \textit{ procedural text describing photosynthesis }, n_s= \textit{ more sunlight }, r_i = \textit{ helps }, h_i = \textit{ 1-hop })$, and $\mathbf{y}_i =  \textit{plants trap sunlight}.$

\ours uses a language model to estimate the probability of generating an end node $n_t$ for an input $\mathbf{x}_i$.
We first transform the 4-tuple $\mathbf{x}_i$ into a single query sequence of tokens by concatenating its components i.e. we set $\mathbf{x}_i = P_i \| n_s \|  r_i \| h_i$, where $\|$ stands for string concatenation.
Let the sequence of tokens representing the target event be $\mathbf{y}_i = \langle y_i^{1}, y_i^{2},\ldots, y_i^{M} \rangle$, where $N$ and $M$ are the lengths of the query and the target event sequences.
We model the conditional probability $p_{\theta}(\mathbf{y}_i \given \mathbf{x}_i)$ as a series of conditional next token distributions parameterized by $\theta$: 
\begin{align*}
    p_{\theta}(\mathbf{y}_i \given \mathbf{x}_i) =  \prod_{k=1}^{M} p_{\theta} (y_i^k \given \mathbf{x}_i, y_i^{1},.., y_i^{k-1}) 
\end{align*}
\ours parameterizes $p_\theta$ using the \textsc{gpt-2}~\cite{radford2019language} pretrained language model.
\textsc{gpt-2} is based on the popular transformer architecture~\cite{vaswani2017attention}, which consists of a series of transformer blocks. 
Each transformer block consists of two operations: a masked version of the multi-headed self-attention~\cite{vaswani2017attention} followed by a feed-forward network (\textsc{ffn}). 
Each of these operations is surrounded by a residual connection~\cite{he2016deep}, and followed by a layer normalization~\cite{ba2016layer} operation.

The auto-regressive factorization of the language model $p_{\theta}$ allows us to efficiently generate target event influences for a given test input $\mathbf{x}_j$.
Each token in $y_j$ is generated by sampling $y_j^{1} \sim p_\theta(y \given \mathbf{x}_j)$.
The next token is then drawn by sampling $y_j^{2} \sim p_\theta(u \given \mathbf{x}_j, y_j^{1})$.
The process is repeated until a specified \emph{end-symbol} token is drawn at the $K^{th}$ step.
The tokens $\langle y_j^{1}, y_j^{2},\ldots, y_j^{K - 1}\rangle$ are then returned as the generated target event influence.

\section{Experiments}

\label{sec:experiments}


\paragraph{Setup:}
We use the dataset described in section \secref{sec:dataset} for training the model. 
The dataset statistics by relation type and hop information are shown in Table~\ref{table:data-split}.
\ours is based on the \textsc{gpt-2} implementation by~\citet{wolf2019huggingface}.\footnote{Details of hyper-parameters in the Appendix~\ref{subsec:appendix_hyperparam}}, and uses nucleus sampling~\citep{holtzman2019curious} for decoding output sequences over the fine-tuned language model.
As discussed in (~\secref{sec:masked-lm}), we concatenate the 4-tuple $\mathbf{x}_i = (P_i, n_s, r_i, h_i)$ in a single sequence of tokens.
$\mathbf{x}_i$ was concatenated using the template:  ``$P$ \textit{what does} $n_s$ $r_i$ \textit{at} $h_i$?''. 
All of our experiments were done on a single Nvidia GeForce RTX 2080 Ti.
The models were fine-tuned for 5 epochs for all the variants.

\subsection{Baselines}
\label{sec:baselines}

\paragraph{LSTM Seq-to-Seq:} We train an \textsc{lstm}~\cite{hochreiter1997long} based sequence to sequence model~\cite{bahdanau2015neural} which uses global attention described in~\cite{luong2015effective}.
We use pre-trained Glove~\cite{pennington2014glove} to initialize the word embedding.\footnote{https://github.com/OpenNMT/OpenNMT-py}

\paragraph{\textsc{gpt-2} w/o Fine-tuning:} The pretrained \textsc{gpt-2}~\cite{radford2019language} without any additional fine-tuning serves as another baseline. 
The goal of this baseline is to understand the extent to which \textsc{gpt-2} without fine-tuning could encode event influence information.\footnote{\textsc{gpt-2} implementation from \citet{wolf2019huggingface}}

\paragraph{COMET:} \comet~\cite{Bosselut2019COMETCT} aims to perform knowledge base completion for commonsense knowledge bases by employing pretrained language models. 
Unlike \ours, \comet~ does not use any context or hop information.
Although \comet's architecture was based on~\textsc{gpt}~\cite{radford2018improving}, our implementation adapts \comet to use GPT-2 for a fair comparison, and also supplement each event input with hop-information. 
More concretely, we set $\mathbf{x}_i = (n_s, r_i, h_i)$, with the goal of generating $\mathbf{y}_i = n_t$. 

\begin{table*}[ht]
\centering
\begin{tabular}{llrrrrrrr}
\toprule
Relation & Hop & BLEU-1 & BLEU-2 & BLEU-3 & BLEU-4 & METEOR & ROUGE & Polarity \\ \midrule
Helps    & 1-hop        &  33.32	& 20.54 &	15.47&	13.16 &	14.88 & 34.19 & 82.56\\
Helps    & 2-hop        &  25.99	&	13.77	&	9.20	&	7.44	&	11.78	&	26.71	&	73.88 \\
Helps    & 3-hop & 32.25 & 17.82 &	14.88 &	13.59 &	14.00 &	33.55 & 80.04 \\
Hurts    & 1-hop & 28.67 & 	16.79 & 	11.71 & 	9.46 & 	13.04 & 	29.37 & 	77.23 \\
Hurts    & 2-hop & 25.03 & 	13.15 & 	8.63 & 	6.74 & 	11.41 & 	26.09 & 	74.12 \\
Hurts    & 3-hop & 32.72 & 	18.08 & 	15.03 & 	13.65 & 	14.46 & 34.19 & 	81.02   \\ \bottomrule
\end{tabular}
\caption{Generation Quality of \ours by Relation Type and Node Distance
}
\label{tab:gen-by-reln}
\end{table*}

\subsection{Automated Evaluation}
For evaluating the predicted event influences, we use the standard evaluation metrics \textsc{bleu}~\cite{papineni2002bleu}, \textsc{meteor}~\cite{denkowski2011meteor}, and \textsc{rouge}~\cite{lin2004rouge}
\footnote{We use~\citet{sharma2017nlgeval} for calculating these metrics. \url{https://github.com/Maluuba/nlg-eval}}.
To complement the above mentioned metrics, we also use \emph{polarity}, which captures the direction of change captured by an influence (increasing, decreasing, neutral).
For example, an event influence ``more sunlight'' has the polarity ``increasing'', whereas an event ``less rain'' has the polarity ``decreasing''.
We calculate the percentage of generated event influences that have the same polarity as the reference.
For example, if both the reference event and the predicted target event are about an `increase,' then we count their polarity to be the same. 
Otherwise, we count their polarity to be different.
We used a small set of hand-curated keywords to detect polarity \footnote{This list of 22 words is included in the Appendix.}. 

Table~\ref{tab:gen-quality} shows that \ours outperforms the baselines on all metrics. 
Furthermore, the results emphasize that the pre-trained models can't generate event influences without being fine-tuned on the task.
\ours outperforms \comet in all the metrics by a considerable margin, (by about 8 ROUGE points), reinforcing the importance of generating knowledge that is grounded in context. 

Table~\ref{tab:gen-by-reln} breaks down the performance of \ours by relation type and the number of hop between the source and the target nodes.
From Table~\ref{tab:gen-by-reln}, we observe that the best performance is obtained on nodes generated at 1-hop with \textit{helps} relation. 
The 1-hop nodes generated with a \textit{hurts} relation perform worse, indicating that generating negative influences is a harder task.
Table~\ref{tab:gen-by-reln} also highlights 
that the 3-hop generations score higher than the 2-hop relations for \textit{help} and \textit{hurts} relations. 
On further inspection, we found that this was an artifact from the human-curated influence graphs. Each influence graph had a maximum hop length of 3, and the end nodes are always of the form ``more \textit{X}'' or ``less \textit{X}'', where \textit{X} is a concept mentioned in the passage.
Due to this templated nature of leaf nodes in the influence graph, the task becomes relatively less challenging compared to 1-hop and 2-hop influences.

\subsection{Human Evaluation}

In addition to automated evaluation, we also compare \ours with \comet for assessing the generation quality using human judgments.
Three human judges annotated 120 unique samples for \textit{relevance} and \textit{reference}, described next.
We also compared the output of the two systems for fluency and found that both the systems produce fluent outputs. This indicates that pretrained language models are effective in generative grammatically correct output, even though the utility of the output may vary greatly.


\begin{table}[h]
\centering
\begin{tabular}{lccc}
\toprule
Task & \ours & No Preference & \textsc{comet} \\
\midrule
Relevance & 46.11 & 30.83 & 23.06 \\
Reference & 31.94 & 56.39 & 11.67 \\
\bottomrule
\end{tabular}
\caption{Results of human evaluation. The numbers show the percentage(\%) of times a particular option was selected for each metric.}
\label{tab:human_eval_relevance}
\end{table}

\begin{table*}[t]
\centering
\resizebox{\textwidth}{!}{%
\begin{tabular}{@{}llllll@{}}
\toprule
 Error Class & Description & \% & Question & Reference & Predicted \\ \midrule
 Polarity & The predicted polarity was wrong  & 5\% & What does `oil fields over-used' & there is not  & more oil \\
 & but event was correct &  & help at 2-hop ? & oil refined & is refined \\ \midrule
 Linguistic & The output was a & 20\% & What does `fewer rabbits will & more  & more \\
 Variability & linguistic variant of the reference &  & become pregnant' hurts at 1-hop ? &  rabbits & babies \\ \midrule
 Related  & The output was related but  & 17\% &  What does you inhale more air  & there will be & you develop  \\
 Event & different reference expected &  & from the outside hurts at 1 hop ? & less oxygen  & more blood clo-  \\ 
 & & & &in your blood& -ts in your veins \\ \midrule
 Wrong & The output was  & 30\% & What does `less nutrients for & more  & more wine  \\
 & was completely unrelated &  & plants' hurt at 2-hop ? & plants & being produced \\ \midrule
 Erroneous  & The gold annotations & 2\% & What does `less rabbit & less  & more  \\
 Reference & were erroneous &  & rabbit mating' hurt at 1-hop? & rabbits & babies \\ 
\bottomrule
\end{tabular}%
}
\caption{Examples of error categories. Error analysis is only shown for the incorrect outputs.}
\vspace{-0.5em}
\label{tab:gen-error-analysis}
\end{table*}

\paragraph{Relevance:} 
The annotators are provided with the 
input of a procedural text, the source event, and the relational questions.
The outputs generated by \comet and \ours are also provided in random order. 
The annotators were asked, ``Which system (A or B) is more accurate relative to the background information given in the context?''
They could also pick option C (no preference).

\paragraph{Comparison with true event (reference):} We measure how accurately each system-generated event reflects the reference (true) event. 
Here, the annotators saw only the reference sentence and the outputs of two systems (A and B) in a randomized order. 
We asked the annotators, ``Which system's output is closest in meaning to the reference?'' 
The annotators could pick the options A, B, or C (no preference). 

For relevance and reference comparison tasks (Table \ref{tab:human_eval_relevance}), we present the percentage of count of human judges for each of the three categories.
The table illustrates that \ours performs better than \textsc{comet} on both the metrics.
Particularly, \ours not only performs better than \comet but also much better than the ``No Preference'' option in the relevance metric.
This means that \ours generates target events that logically follow the passage and source events.
We note that the automated metrics may not capture the relevance and correctness of the generated target events.
The reference and relevance task scores together show that \ours does not generate target events that are exactly similar to the reference target events, but they are correct in the context of the passage and the source event.
This can happen due to linguistic variation in the generation, as well as the ability of the source event to influence multiple target events in the context of the passage.
We study this in more detail in the error analysis presented below.

\subsection{Error Analysis}
\label{subsec:error-analysis}
Table~\ref{tab:gen-error-analysis} shows the error analysis on 100 random samples from the validation set.
We found that for about 26\% of samples, the generated event influence had an exact match with the reference, and about 30\% of the samples had no overlap with the reference (category \emph{Wrong} in Table \ref{tab:gen-error-analysis}). 
We found that for 20\% of the cases, the generated target event was correct but was expressed differently compared to the reference text (\emph{Linguistic Variability}) class in Table \ref{tab:gen-error-analysis}).
Furthermore, we observed that in 17\% of cases, the generated target event was not the same as the reference target event, but it was relevant to the passage and the question, as shown in the \emph{Related Event} category in Table~\ref{tab:gen-error-analysis}.
In 5\% of the samples (\emph{Polarity}), the model generates events with opposite polarity compared to the reference. 
A small fraction (2\%) of samples had incorrect gold annotations.

\subsection{Ablations and Discussion}
Table~\ref{tab:ablation} shows the ablation results by removing each of paragraph, reverse edges and hop information from the 4-tuple $\mathbf{x}_i = (P_i, n_s, r_i, h_i)$~( \secref{sec:task}).
These ablations are performed to get an insight into the contribution of each component in the input $\mathbf{x}_i$ to the generation task.
In line with the expectation, the model with access to all the input components performs the best on almost all of our evaluation metrics.
We also observe that the context information was the best indicator of model performance gains.
Our ablation results re-emphasizes that grounding event influence in the context of the passage is crucial for the generation of target events.

\begin{table*}[!htbp]
\centering
\begin{tabular}{lllrrrrrrr}
\toprule
Para   & Rev   & Hop & BLEU-1    & BLEU-2   & BLEU-3   & BLEU-4   & METEOR & ROUGE & Polarity \\ \midrule
\xmark & \xmark & \cmark &   20.20 &	9.48&	6.63&	5.46&	8.83&	20.49&	73.75 \\
\xmark & \cmark & \xmark &  19.96 & 9.19 & 6.49 & 5.41 & 8.58 & 20.39 & 72.80 \\
\xmark & \cmark & \cmark & 20.63&	10.01&	7.12&	5.93&	8.28&	20.93&	71.97 \\
\cmark & \xmark & \xmark &  26.19&	14.08&	10.06&	8.52&	11.85&	26.78&	74.67  \\
\cmark & \xmark & \cmark & 27.51 &	15.64 &	11.68 &	10.02 &	12.42 &	27.81 &	76.76 \\
\cmark & \cmark & \xmark & 26.05& 14.23&	10.10 & 8.45 & 11.87 & 27.10 & 75.68 \\
\cmark & \cmark & \cmark & 28.97 &	16.23 &	11.69&	9.74&12.85&	29.65 &77.24
  \\ \bottomrule
\end{tabular}
\caption{Ablation experiments to understand the contribution of each of paragraph, reverse edges and hop information to the generation of the target event.}
\vspace{-0.35em}
\label{tab:ablation}
\end{table*}

\section{Downstream QA} 

In this section, we examine the utility of \ours-generated graphs in a downstream question answering task on the \wiqa benchmark.

\subsection{The QA Task}
\wiqa~\cite{tandon2019wiqa} is a dataset for procedural understanding, that comprises of ``what-if" questions to reason about the effects of one event perturbation on another in the context of a process.
Specifically, each question in \wiqa consists of a context paragraph $P$ and two input events $n_c$ and $n_e$.
The task is to predict how $n_c$ affects $n_e$, where the result is one of: \{\textit{helps}, \textit{hurts}, and \textit{no\_effect}\}.



\subsection{Using \ours to augment \qa data}
We use \ours to augment the event influences in each sample in the \qa task as additional context.
Concretely, for the given context $P$, and the event influences $n_c$ and $n_e$, we generate forward influences for $n_c$ and reverse influences for $n_e$ using \ours.
This scheme is intended to generate reasoning chains that connect $n_c$ to $n_e$, even if $n_e$ is not an immediate consequence of $n_c$.
Concretely, we query \ours with four inputs: $(P, n_c, \textit{helps}, \textit{1-hop})$, $(P, n_c, \textit{hurts}, \textit{1-hop})$, and $(P, n_e, \textit{helped by}, \textit{1-hop})$, $(P, n_e, \textit{hurt by}, \textit{1-hop})$.
The generated event influences are then concatenated to form a flattened list of sentences $x_g$.

Following \citet{tandon2019wiqa}, we encode the input sequence $P \| n_c \| n_e $ using the \textsc{bert} encoder $E$~\cite{devlin-etal-2019-bert}, and use the \textsc{[cls]} token representation ($\mathbf{\hat{h_i}}$) as our sequence representation.
We then use the same encoder $E$ to encode the generated influences $x_g \| n_c \| n_e$, and use the \textsc{[cls]} token to get a representation for augmented influences ($\mathbf{\hat{h_a}}$).
Following the encoded inputs, we compute the final loss as follows: 
\begin{align*}
    \mathbf{l_i} &= \texttt{MLP\textsubscript{1}}(\mathbf{\hat{h_i}}) \\  \nonumber
    \mathbf{l_a} &= \texttt{MLP\textsubscript{2}}(\mathbf{\hat{h_a}}) \\  \nonumber
    \mathcal{L} &= \alpha \times \mathcal{L}_i + \beta \times \mathcal{L}_a 
\end{align*}

where $\mathbf{l_i}, \mathbf{l_a}$ represent the logits from $\mathbf{\hat{h_i}}$ and $\mathbf{\hat{h_a}}$ respectively, and $\mathcal{L}_i$ and $\mathcal{L}_a$ are their corresponding cross-entropy losses. $\alpha$ and $\beta$ are hyperparameters that decide the contribution of the generated influence graphs and the procedural text to the loss. For our experiments, we set $\alpha = 1$ and $\beta = 0.9$. 

\subsection{QA Evaluation Results}
Tables~\ref{tab:qa-acc}, \ref{tab:accuracy-hops}, and \ref{tab:acc-question-type} show the results from our experiments on the \wiqa QA dataset.
\bert refers to the results from the original \bert based implementation by \citet{tandon2019wiqa}, and \bert + \ours are the results obtained by augmenting the QA dataset with the influences generated by \ours as described above.
Further, Tables~\ref{tab:accuracy-hops} and \ref{tab:acc-question-type} show the accuracy of our method vs. the vanilla \bert model by question type and number of hops between $n_c$ and $n_e$.
We observe from Table~\ref{tab:accuracy-hops} that augmenting the context with generated influences from \ours leads to considerable gains over \bert based model, with the largest improvement seen in 3-hop questions.
The strong performance on the 3-hop question supports our hypothesis that generated influences might be able to connect two event influences that are farther apart in the reasoning chain.
We also show in Table~\ref{tab:acc-question-type} that augmenting with \ours improves performance on the difficult exogenous category of questions, which requires background knowledge.
In summary, the evaluation highlights the value of \ours as a framework for improving performance on downstream tasks that require event-based background reasoning and serves as an evaluation of the ability of \ours to generate targeted influences. 

\begin{table}[t]
\centering
\begin{tabular}{lr}
\toprule
Model     &  Accuracy \\ \midrule
\bert + \ours    &  \textbf{76.92}   \\
\bert    &  73.80  \\
\bottomrule      
\end{tabular}
\caption{QA Accuracy 
}
\label{tab:qa-acc}
\end{table}

\begin{table}[t]
\centering
\begin{tabular}{ccc} 
\toprule
Query Type &       \bert + \ours         & \bert       \\ 
\midrule
1-hop      & \textbf{78.78}  & 71.60  \\
2-hop      & \textbf{63.49}          & 62.50  \\
3-hop      & \textbf{68.28}  & 59.50  \\
\bottomrule
\end{tabular}
\caption{QA accuracy by number of hops}
\vspace{-1em}
\label{tab:accuracy-hops}
\end{table}


\begin{table}[t]
\centering
\begin{tabular}{lrr}
\toprule
Question Type   & \bert + \ours  & \bert \\ \midrule
Exogenous  & \textbf{64.04}  & 56.13        \\
In-para &     73.58 & \textbf{79.68}     \\
Out-of-para & \textbf{90.84} &89.38 \\ \bottomrule      
\end{tabular}
\caption{QA accuracy by question type}
\vspace{-1em}
\label{tab:acc-question-type}
\end{table}

\section{Conclusion}
\label{sec:conclusion}

We define the problem of event-influence reasoning as a generation task conditioned on context, particularly exploring the efficacy of large scale pre-trained language models for the task.
We use human-curated event influence graphs to train a model to generate targeted event influences grounded in a context. 
Our experiments with ablations and error analysis provide insights into how to effectively adapt pretrained language models for event influence generation and opens up exciting avenues for further research.
Our method outperforms strong baselines on both automated and human evaluations.
Furthermore, generated influences improve performance on the benchmark \wiqa \qa task without architectural changes to the model.
Future work would extend the generalizability of this method to understand more complex and volatile event influences, such as events in news articles and stock markets.
\section*{Acknowledgments}
This material is based on research sponsored in part by the Air Force Research Laboratory under agreement number FA8750-19-2-0200, and in part by grants from  National Science Foundation Secure and Trustworthy Computing program (CNS-1330596, CNS-15-13957, CNS-1801316, CNS-1914486). 
The U.S. Government is authorized to reproduce and distribute reprints for Governmental purposes notwithstanding any copyright notation thereon. 
The views and conclusions contained herein are those of the authors and should not be interpreted as necessarily representing the official policies or endorsements, either expressed or implied, of the Air Force Research Laboratory, the NSF, or the U.S. Government.

\bibliography{main}
\bibliographystyle{acl_natbib}
\clearpage
\appendix
\section{Appendix}
\subsection{Polarity Words}

\paragraph{Increasing words} \textit{helps, more, higher, increase, increases, stronger, faster, greater, longer, larger, helping}
\paragraph{Decreasing words} \textit{hurts, less, lower, decrease, decreases, weaker, slower, smaller, hurting, softer, fewer}

\subsection{Hyperparameters}
\label{subsec:appendix_hyperparam}

\paragraph{Seq-to-Seq:} We use 2 layers of LSTM encoder and decoder with a hidden size of 500, word embedding size of 300. The encoder is bidirectional. We use Glove embedding of 300 dimensions.

\paragraph{\ours:}
\ours fine-tunes  \textsc{gpt-2}~\citep{radford2019language}, allowing us to re-use the same hyperparameters as with small adjustments in the recommended range.
We use the medium (355M) variant of \textsc{gpt-2} for our experiments with 24-layer, 1024-hidden, 16-heads, 345M parameters (\url{https://huggingface.co/transformers/pretrained_models.html}). 
We use the weights released by~\citet{radford2019language}. 
We use Adam~\cite{kingma2014adam} for optimization with a learning rate of $5e-05$.
All the dropouts~\cite{srivastava2014dropout} were set to 0.1
We found the best hyperparameter settings by searching the space using the following hyperparameters. 
\begin{enumerate}
    \item weight decay = \{ 0.1, 0.01, 0.05 \}
    \item embedding dropout = \{0.1, 0.2, 0.3 \}
    \item learning rate = \{1e-05, 2e-05, 5e-05, 1e-06\}
\end{enumerate}

\end{document}